\documentclass[a4paper]{article}
\usepackage{diagbox}
\usepackage{spconf,amsmath,graphicx}
\usepackage{xcolor}
\usepackage{array}

\title{ENABLING ZERO-SHOT MULTILINGUAL SPOKEN LANGUAGE TRANSLATION WITH\\ LANGUAGE-SPECIFIC ENCODERS AND DECODERS}

\name{Carlos Escolano, Marta R. Costa-jussà, José A. R. Fonollosa, Carlos Segura$^*$}
\address{TALP Research Center, Universitat Politècnica de Catalunya, Barcelona\\ $^*$Telefónica I+D, Barcelona}
\begin{document}

\maketitle
\begin{abstract}
Current end-to-end approaches to Spoken Language Translation (SLT) rely on limited training resources, especially for multilingual settings. On the other hand, Multilingual Neural Machine Translation (MultiNMT) approaches rely on higher-quality and more massive data sets.
Our proposed method extends a MultiNMT architecture based on language-specific encoders-decoders to the task of Multilingual SLT (MultiSLT). Our method entirely eliminates the dependency from MultiSLT data and it is able to translate while training only on ASR and MultiNMT data.

Our experiments on four different languages show that coupling the speech encoder to the MultiNMT architecture produces similar quality translations compared to a bilingual baseline ($\pm 0.2$ BLEU) while effectively allowing for zero-shot MultiSLT. Additionally, we propose using an Adapter module for coupling the speech inputs. This Adapter module produces consistent improvements up to +6 BLEU points on the proposed architecture and +1 BLEU point on the end-to-end baseline.
\end{abstract}


\section{Introduction}

Research in end-to-end Spoken Language Translation (SLT) has appeared as an effective alternative to the classical cascade approach, i.e. the concatenation of Automatic Speech Recognition (ASR) plus Machine Translation (MT) systems. The advantages of the end-to-end systems are mainly accounting for a significant reduction of inference time; direct use of prosodic features; and more relevantly, the avoidance of concatenation errors. 
Even though in recent years new datasets have been released \cite{DBLP:conf/lrec/WangPWG20, di-gangi-etal-2019-enhancing}, the available data for SLT is still limited, compared to cascade approaches that rely on ASR and MT datasets. Especially the latter, that can usually be orders of magnitude larger. In parallel, Multilingual Neural Machine Translation (MultiNMT) systems have been improving the quality of translation, mostly for low-resource pairs that learn from high-resource ones, even allowing the translation of language pairs never seen in training, zero-shot translation. These architectures can vary the degree of shared information between the different languages, from universal models where a single model processes all languages \cite{johnson2017google} to completely language-specific parameters relying on the cross-lingual mapping of several languages into a shared representation space \cite{escolano:2020}. Other works rely on Multilingual SLT data \cite{9003832, li2021multilingual} to train the system, which face the problem of the scarcity of resources.

 In this work, we propose an end-to-end model that can perform SLT to all the languages supported by a given MultiNMT model by just training one translation direction or ASR task. This is done efficiently by only training a new speech encoder and benefiting from all the data sources available, ASR, SLT, and MultiNMT. To do so, we propose to couple a speech encoder to the MultiNMT architecture based on language-specific encoders-decoders. This coupling extends the previous MultiNMT system to a new data modality. It allows the system to translate between several languages benefiting from the abundance of  MultiNMT data available in terms of quantity and number of languages.
 We only require parallel data from speech utterances in one language and one of the languages supported by the MultiNMT system, including ASR data on the same language. Once trained, we can account for zero-shot Multilingual SLT (MultiSLT) between the added spoken language to any other languages already in the system, without any explicit training for the task or parallel data for those directions.  %Without parallel data for those directions, zero-shot MultiSLT.
 
 The two main contributions of this paper are: (1) The training of a zero-shot MultiSLT system using a trained MultiNMT system, both from bilingual SLT and monolingual ASR corpora, without the need of MultiSLT corpora (2) The use of an Adapter module \cite{bapna2019simple} for SLT, which leads to significant improvements, up to +6 BLEU points on the proposed MultiSLT architecture on the supervised directions, up to +3 BLEU points on zero-shot, and  +1 BLEU points in performance on the end-to-end baseline systems. Additional discussion provides further study of the impact of the Adapter on the models hidden representation.
 
 %{\color{red} In order to contextualize our results we also provide a discussion of the results paying attention the differences between the representations learnt from both textual and speech data modalities and the impact of the Adapter modules in such space.}
 
 %{\color{red} The paper is organized as follows: Background and related methods \ref{sec:background}, proposed methodology \ref{sec:propo}, Experimental Framework\ref{sec:framework} with the data and implementation details of the proposed experiments, Results\ref{sec:results} on the task focusing on the tasks of SLT and Zero-shot SLT, Discussion\ref{sec:discussion} on the impact on the cross-lingual mapping between languages and modalities and Conclusions\ref{sec:conclusions}.}

\section{Background}
\label{sec:background}

In this section, we describe the previous work that is used as starting point of our proposal, which are the SLT architecture of the S-Transformer \cite{di-gangi-etal-2019-enhancing} and the MultiNMT architecture of language-specific encoder-decoders \cite{escolano:2020}. 

\subsection{S-Transformer for SLT}
\label{sec:stransf}

The S-Transformer \cite{di-gangi-etal-2019-enhancing} presents a specific adaptation of the Transformer \cite{vaswani2017attention}, %, which uses 2D self-attention \cite{dong:2018}, residual connections and a distance penalty in the Transformer layers to attend near samples \cite{sperber:2018}. 
  which is based on 3 main modifications of its encoder, while keeping the same decoder. First, it downsamples the speech input, which is much longer than text inputs, with convolutional neural networks \cite{dong:2018}. Second, it models the bidimensional nature of the audio spectrogram with two-dimensional (2D) components. Finally, it adds a distance penalty to the attention creating an attention bias towards short-range dependencies \cite{sperber:2018}. This S-Transformer has been shown to outperform previous baseline systems both in translation quality and computational efficiency and has established a new state-of-the-art performance on end-to-end SLT for six language pairs. Previous works \cite{9004003} have shown that this architecture is extendable to a multilingual setting, but it has, to our knowledge, never been used on the zero-shot scenario.

\subsection{Language-Specific Encoders and Decoders for Multi\-NMT}
\label{sec:langspe}

This approach \cite{escolano:2020} trains a separate encoder and decoder for each of the $N$ languages available without requiring multi-parallel corpus but requiring parallel corpus among all ${N(N-1)}$ translation directions. This approach jointly trains independent encoders and decoders for each translation direction. In this joint training, the main difference from standard pairwise training is that, in this case, there is only one encoder and decoder for each language, which is used for all translation directions involving that language. The approach does not share any parameter across modules from different languages. Training all language-specific modules to perform the different translations tasks enforces the system to converge to a representation space shared between all encoders and decoders. This means that this architecture can be incrementally extended to new modules (other languages or modalities) by enforcing the new module to learn this shared representation. This architecture does not require retraining the entire system when adding new languages and it requires only parallel data from the new language or modality to one of the languages already in the system. Therefore, this is allowing for zero-shot translation from the added language and the other $N-1$ languages in the system. This approach has been shown to outperform the MultiNMT shared encoder/decoder architecture \cite{escolano:2020}.

\section{Proposed Methodology}
\label{sec:propo}

In this work we extend the language specific encoder-decoder architecture to spoken modules. For this purpose, we adapt the S-Transformer to be compatible with the language specific encoder-decoder architecture and therefore accounting for zero-shot MultiSLT.

\begin{figure*}[t]
\begin{center}
\includegraphics[scale=0.2]{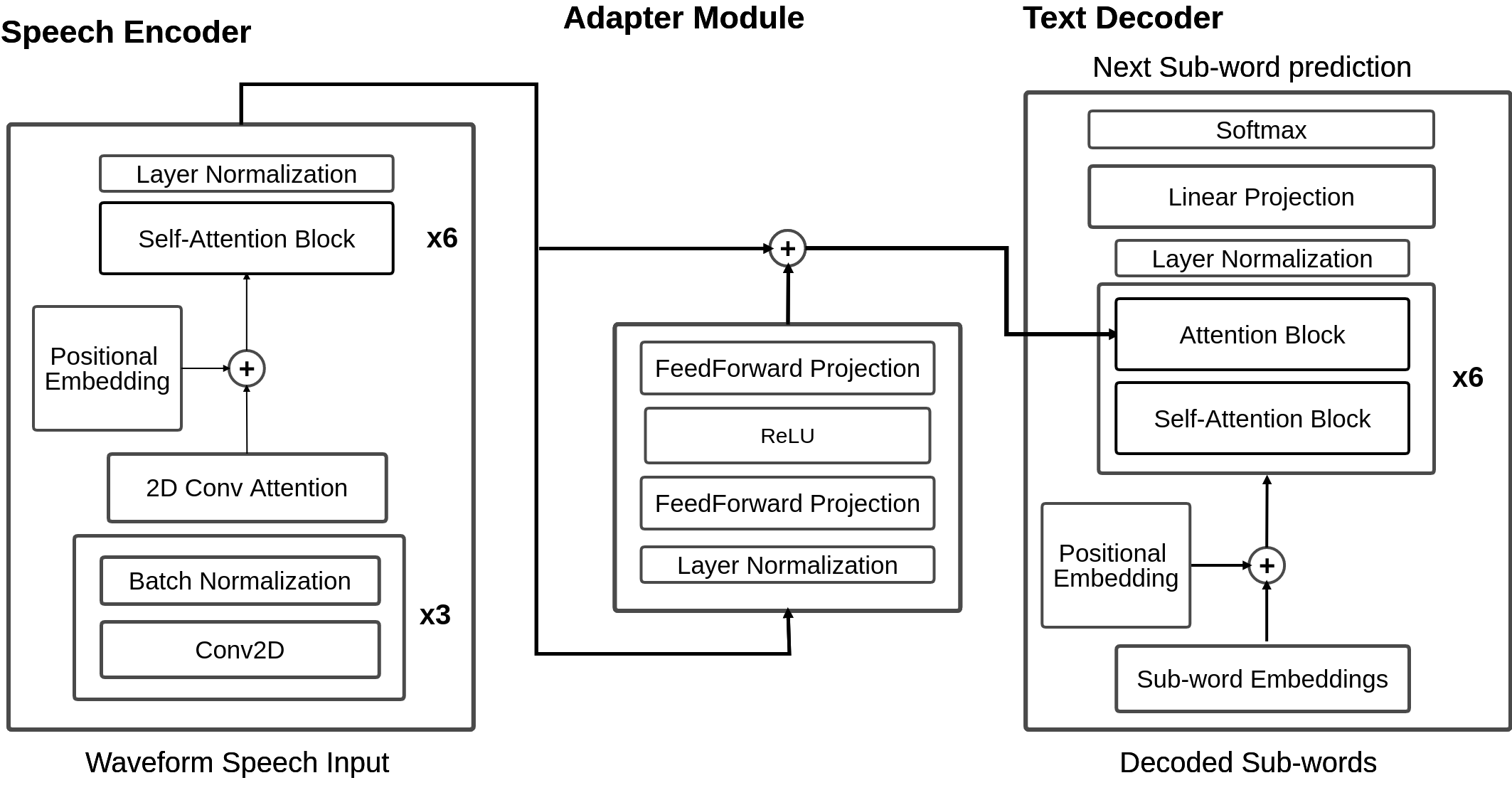}
\caption{Close detail to our proposed architecture and its main components, speech encoder, Adapter and text decoder. Dotted square shows the frozen parts of the architecture during training.}
\label{fig:architecture}
\end{center}
\vspace{-0.75cm}
\end{figure*}

\subsection{Speech Encoder}

There are several adaptations that the speech encoder based on the S-Transformer described in Section \ref{sec:stransf} requires to be compatible with the already existing language-specific encoder-decoders described in Section \ref{sec:langspe}. First, we use an extra 2D-convolutional layer, in order to further reduce the length of the input speech, resulting in a total of 3 2D-convolutional layers, as in previous works \cite{hannun:2019}. Each convolution, with kernel size 3 and stride 2, halves the temporal dimension of the input waveform resulting in an 8 times shorter sequence, obtaining sequences 2 times longer than the text target on average, measured on the validation set. Second, we make use of an Adapter module at the end of the speech encoder. This modeling technique has been previously proposed for NLP \cite{bapna2019simple,artetxe2019cross} and used in a variety of tasks including multilingual ASR with a single model \cite{google:2019}. Previous works use this technique to finetune a frozen model for specific languages or tasks. They provide a lightweight mechanism to adapt representations to new languages of tasks by only training a small portion of the model's total number of parameters.  In our case, we extend this functionality to bridge the differences between speech and textual modalities. We start from a pretrained encoder and decoder and using an Adapter module during training allows to better learn the difference in the intermediate representation created by the speech encoder and the language-specific encoder-decoders, helping the model to overcome multimodal differences.

The Adapter module consists of a step of layer normalization followed by a projection step with ReLU non linear activation. This step projects the encoder's hidden representation into a different dimensionality space.  This representation is projected back to the original dimensionality by a second feedforward layer. This process tries to capture new information in a feature richer space while trying to maintain that information when recovering the original dimensionality. The final encoder representation will be the sum of the self-attention encoder and adapter outputs, as a residual connection. Being the projection dimensionality the only hyperparameter of the module to tune allows a fast tuning for each language. See Figure \ref{fig:architecture} for a block diagram of the proposed architecture.

\subsection{Adding Speech Modules to the MultiNMT architecture based on Language-Specific Encoders and Decoders} 

Since parameters are not shared between the independent encoders and decoders, the joint training enables the addition of new languages and modalities without the need to retrain the existing modules. Following the method proposed by \cite{escolano-etal-2019-multilingual} speech encoders are added by training them with a frozen decoder from the MultiNMT system.
Let us say we want to add speech of the language $N$ (hereinafter, $N_{speech}$). To do so, we must have parallel data between $N_{speech}$ and any language in the system. For illustration, let us assume that we have $N_{speech}-I$ parallel data. 
Then, we can set up a new bilingual system with language $N_{speech}$ as source and language $I$ as target. To ensure that the representation produced by this new pair is compatible with the previously jointly trained system, we use the previous $I$ decoder ($d_{I}$) as the decoder of the new $N_{speech}$-$I$ system and we freeze it.  $N_{speech}$ encoder is a speech decoder following the architecture described before, pretrained on ASR data with the exception of the Adapter module that is randomly initialized. During training, we optimize the cross-entropy between the generated tokens and  $I$ reference data but update only the parameters of the $N_{speech}$ encoder ($e_{{N_{speech}}}$) and the Adapter module. By doing so, we train $e_{{N_{speech}}}$ not only to produce good quality translations but also to produce similar representations to the already trained languages. Figure \ref{fig:architecture} illustrates this process and how the modules interact during training, with the frozen text decoder for language $I$ and the new modules for $N_{speech}$.

\section{Experimental Framework}
\label{sec:framework}

Experiments are run as follows. We build a baseline system which consists of an end-to-end ASR and SLT architecture based on the S-Transformer \cite{di-gangi-etal-2019-enhancing}. %The ASR English encoder is used as pre-trained encoder for the rest of SLT experiments.
Using the pre-trained ASR English encoder, we train the SLT systems from English-to-German, French and Spanish (Baseline); we then alternatively add our proposed architecture presented in Section \ref{sec:propo} of the language-specific architecture (LangSpec) or the Adapter module (Baseline \& Adapt). Finally, we add the combination of both proposed architectures (+LangSpec \& Adapt). As further comparison we present cascade results (Cascade) from the ASR model trained for each direction and the language-specific MultiNMT. ASR results are provided in Word Error Rate (WER), while SLT results are provided in Bilingual Evaluation Understudy (BLEU) \cite{papineni-etal-2002-bleu}.

\subsection{Data}

The MultiNMT system of language-specific encoder-decoders is trained on the EuroParl corpus \cite{Europarl}, consisting of speech transcriptions from the European Parliament, in English to German, French, Spanish as training data, with 2 million parallel sentences among all combinations of these four
languages (without being multi-parallel). As validation and test set, we used \textit{newstest2012/2013}, respectively, from WMT13 \cite{bojar-EtAl:2013:WMT}, which is multi-parallel across all the above languages. All data were preprocessed by applying tokenization, punctuation normalization and truecasing by using standard Moses scripts \cite{koehn2007moses}. Finally, data was tokenized into subwords by applying BPE \cite{sennrich-etal-2016-neural} with 32k operations using subword-nmt library. \footnote{https://github.com/rsennrich/subword-nmt}

We use Must-C as speech dataset \cite{di-gangi-etal-2019-enhancing}, which is a multilingual set extracted from TED talks in English with transcriptions and textual translations in 8 languages. Must-C is the largest corpus in the languages that we are aiming at (English, German, French, Spanish). The amount of transcribed hours varies from 385 to 504, depending on the language pair. We are using the English transcribed speeches and the bilingual data on pairs English-German, English-French, English-Spanish. We use the training, validation and test splits that Must-C provides. Multilingual validation and test sets have around 1.4K and 2.5K sentences (respectively) varying on the language pair. For all sets the provided speech preprocessing is used, 40 dimensional log Mel spectrograms computed with windows of 25ms and hop length of 10ms. For the textual part the same preprocess used for the MultiMT data is applied, using the same vocabularies and BPE codes. 

Note that speech and text data are from different domains, which may lead to differences in the vocabulary and general writing style of the texts. 
Studying the effect on this matter is out-of-scope.

\subsection{Parameters}

Our MultiNMT system is based on language-specific encoder-decoders, and we use the same architecture proposed in \cite{escolano:2020}. This architecture uses the Transformer with 6 attention blocks for both encoder and decoder and 8 attention-heads each and 512 hidden dimensions, with the only modification of adding layer normalization as the last step of both encoder and decoder.  Experimental results showed adding this step to the original textual space helped model convergence during the SLT training. Our baseline SLT architecture follows the parameters  in \cite{di-gangi-etal-2019-enhancing}. The basic parameters are identical to the MultiNMT system, with the only modifications to the original system of using 3 2D-convolutional layers (instead of 2) and 2048  hidden dimensions for the feedforward layers to match the MultiNMT model.  The addition of a 3rd 2D-convolutional helped the model learn the mapping to the pre-trained space. On average, over the validation set, this layer reduced the input speech sequence from 4 times longer to just 2 times longer than text, which can help trained decoder attend the input.

\begin{figure}[t!]
\begin{center}
\includegraphics[scale=0.63]{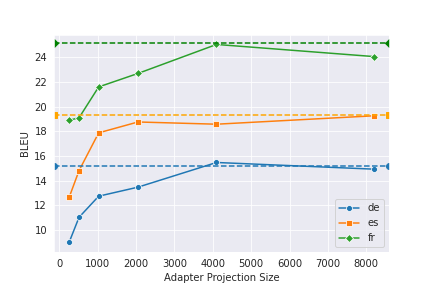}
\caption{Model performance for several values of the Adapter projection size for Spanish, German and French. The baseline model in dotted lines}
\label{adapter-size}
\end{center}
\vspace{-0.75cm}
\end{figure}

Finally, we tested several alternatives to projection size for the Adapter module to see their effect on the task. %Several works have proposed a projection to a lower dimensionality space to force an information bottleneck and over parametrized the space to capture new information about the data. 
Figure \ref{adapter-size} shows the performance of the model for the tasks of English to Spanish, French, and German SLT compared to each respective bilingual baseline system. 
All three tasks projecting the encodings to a lower dimensionality space were harmful to the task, even when keeping the original 512 dimensions. 
When over-parametrizing the space, we observe an improvement in performance in all cases, especially for 4096 dimensions. We tested our models until 9120 dimensions as it was the biggest size we could use without out of memory errors on a single GPU. Experiments show that the models obtain their best performance (close to or slightly better than the baseline system) at 4096  dimensions for German and French and 9120 dimensions for Spanish and English using ASR data. We also measured the impact on the number of parameters of the model by the addition of the Adapter modules. For all languages, baseline models have approximately 77 million parameters with slight differences due to each language vocabulary. Adapters with 4096 projection size accounted for additional 4 millions parameters, or 5,5\% of the total number of parameters. 9192 projections size accounted for 8 million parameters, or 11\% of the total number of parameters. This numbers show that Adapter modules are a lightweight and easy to tune option for this task.

\section{Results}
\label{sec:results}

As our proposed approach is able to leverage ASR, SLT, and MultiNMT data, we propose two different baseline architectures to compare the impact of these data sources. These two systems are a cascade approach (Cascade) and an end-to-end SLT baseline (Baseline). The cascade approach uses MultiNMT to translate the output of an ASR model. Both models were trained individually using MultiNMT and ASR data, respectively. The end-to-end baseline approach (Baseline) uses the S-Transformer architecture, trained with SLT data.

Table \ref{results-de} shows that adding a new speech encoder to the MultiNMT language-specific architecture (LangSpec) is possible by using the existing architecture of the S-Transformer encoder and supervised training with one of the languages in the system (English (en), German (de), French (fr), Spanish (es)). We are training a new English speech encoder with a frozen target module of the MultiNMT language-specific architecture (either en, de, fr, or es). BLEU results vary from 10.80 in English-to-German up to 19.10 in English-to-French. When compared to the baseline systems, we observe a gap of +5 BLEU points due to the difference between modalities and not training the decoder for the task of SLT. These results are improved in a large amount when adding the Adapter module (\& Adapt), up to almost 6 points BLEU (English-to-French), reducing the gap to the end-to-end baseline system to $\pm 0.2$ BLEU for all tested SLT directions. In fact, the Adapter module by itself consistently improves in all directions on SLT models, even when applied to the Baseline, that is trained from scratch, by +1 BLEU points. This shows that the Adapter is helpful to bridge the modalities' representation in all scenarios, not only when limiting the parts of the network trained and both encoder and decoder are jointly trained for the task of SLT. Note that the LangSpec coupling is allowing for zero-shot translation with other languages available in the MultiNMT system  independently of the availability of SLT data for those directions. We test this behaviour by translating to all 3 languages using the same models that were trained only on one of those directions or ASR. Table \ref{zero-shot} shows the results in zero-shot (except for the numbers in italics that show supervised results) with and without Adapt. Zero-shot is achieved whenever we train the English speech encoder with any text decoder available in the LangSpec architecture (en, de, fr, es). For example, column SLT$_{ende}$ is showing results for our model trained only on the English to German SLT task on English to German, French, and Spanish without additional training. In this case, we are training the English speech encoder with the German text decoder (SLT$_{ende}$  model), and we have zero-shot with English speech to French (tgt fr) and Spanish (tgt es). Adapter modules consistently improves in all translation directions on SLT models, reducing the gap between end-to-end systems and cascade. We also observe that higher zero-shot results are achieved when training on the French text decoder which is consistent with the supervised results, indicating that the selection of the supervised language pair may have an impact on the final performance of the zero-shot translation directions, and that higher results on the supervised directions may be related to better mappings between the two data modalities.

In addition to models trained with SLT data, we also performed the same zero-shot experiments with a model trained using only ASR data in English (ASR$_{en}$). Results show that zero-shot translation can also be achieved in this scenario for all the other tested languages and the quality of the results correlates with the supervised SLT performance, being French the best direction at 10.85 BLEU points. By contrast, unlikely the models trained with SLT data, the addition of adapters seems to harm the zero-shot performance when added to the ASR$_{en}$, despite showing a small improvement of 0.23 of WER on the supervised ASR task. This discrepancy could be explained by the differences in the tasks (e.g. monotonic vs non-monotonic alignment) or the less semantic nature of the task of ASR, compared to SLT.

\begin{table} [t!]
\centering
\small
\begin{tabular}{|m{1.4cm}|l|l|l|l|}
\hline System & ASR$_{en}$ & SLT{$_{ende}$} & SLT{$_{enfr}$}  & SLT{$_{enes}$} \ \\ \hline
Cascade & - & 17.30 & 27.15 & 21.29\\ \hline
Baseline & 28.75 & 15.19 & 25.18 & 19.36\\ \hline
Baseline \& Adapt & 28.54 & 16.40  & 26.87 & 20.90 \\ \hline
LangSpec & 29.60 & 10.80  & 19.10 & 14.23\\ \hline
LangSpec \& Adapt & 29.37 & 15.48 & 25.04 & 19.26 \\
\hline
\end{tabular}
\caption{\label{results-de} WER results for ASR$_{en}$  and BLEU results for SLT{$_{ende}$}, SLT{$_{enfr}$}, SLT{$_{enes}$} models, each trained only its specific task.}
\end{table}

\begin{table} [t!]
\centering
\small
\begin{tabular}{|l|l|l|l|l|}
\hline 
% \toprule
\backslashbox{Tgt}{Model}  & ASR$_{en}$ & SLT{$_{ende}$} & SLT{$_{enfr}$}  & SLT{$_{enes}$} \\ \hline
%\multicolumn{1}{|l|}{en}   & \underline{\textit{29.60}} & 66.62 & 66.77 & 64.13\\ 
% \multicolumn{1}{|r|}{+Adapt}& \underline{\textit{29.37}} & 59.25 & 60.49 & 66.83 \\ \hline
\multicolumn{1}{|l|}{de}   & 6.77 & \underline{\textit{10.80}} & 8.43 & 8.05\\ %\cline{2-5}
 \multicolumn{1}{|r|}{+Adapt}& 5.58 & \underline{\textit{15.48}} & 9.41  &  8.13\\ \hline
\multicolumn{1}{|l|}{fr}   & 10.85 & 10.66  & \underline{\textit{19.10}}  & 14.06 \\ %\cline{2-5}
 \multicolumn{1}{|r|}{+Adapt}& 8.27  & 13.48 & \underline{\textit{25.04}}&  14.46\\ \hline
 \multicolumn{1}{|l|}{es}   & 6.75 & 8.14 & 9.83 & \underline{\textit{14.23}} \\ %\cline{2-5}
 \multicolumn{1}{|r|}{+Adapt}& 5.64 & 10.61 & 11.25 & \underline{\textit{19.26}}\\ 
\hline
\end{tabular}
\caption{\label{zero-shot} Zero-shot BLEU results from English to different targets (Tgt) (German (de), French (fr), Spanish (es)) using 4 supervised models (ASR$_{en}$, SLT{$_{ende}$}, SLT{$_{enfr}$}, SLT{$_{enes}$}). Supervised results are in italics and underlined.}
\end{table}

\section{Discussion}
\label{sec:discussion}

After studying the performance of our method, one question that arises is how similar the obtained speech representations are to the textual ones from our MultiNMT data. Zero-shot performance would benefit from mapping all languages and modalities in the same space. The newly learned representation would be more similar to the ones the system was trained with. This mapping of two data modalities into a shared space becomes more challenging than mapping different languages due to the different natures of the data. Speech utterances may have even an order of magnitude more elements than their textual transcription/translations and are split in arbitrarily given their sampling frequency.

\begin{figure}[t!]
\begin{center}
\includegraphics[scale=0.66]{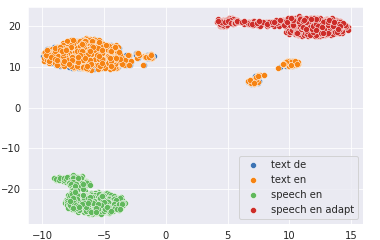}
\caption{Visualization. Speech representation without Adapter (green, bottom left), with Adapter (red, top right) and text representation English and German, which are overlapped, (orange,blue respectively, top left)}
\label{visualization}
\end{center}
\vspace{-0.75cm}
\end{figure}

\textit{Visualization of the intermediate representation.} We further analyze our model by providing a visualization of the intermediate representation and reporting the accuracy in cross-modal sentence retrieval. We use a tool \cite{escolano-etal-2019-multilingual} freely available\footnote{https://github.com/elorala/interlingua-visualization} that allows to visualize in the same space the intermediate sentences representations from different languages. The tool uses the encoder output fixed-representations as input data and makes a dimensionality reduction of these data using UMAP \cite{mcinnes2018umap-software} to visualize the sentence representations into a 2D plot. We compare the intermediate representation for the speech and text segments for 820 sentences (randomly extracted from the Must-C test set) in Figure \ref{visualization}. We observe that the speech representations are far from the text representations, which are altogether in the same part of the space (English/German text overlap). Instead of projecting the speech representation into the region where text representation is found, the Adapter module seems to provide additional information to the representation, but it does not create a  mapping between modalities. The distance between models is only reduced by a small amount and the distribution of the tokens in the space is similar in both cases, indicating that the relative distance between sentences from the same set is preserved.

\textit{Cross-modal sentence retrieval.} If our hypothesis is correct, the relative position of the sentence in the space should be similar with or without the Adapter. Showing a clear correspondence between them With the same set, we performed a top-1 sentence retrieval computing the cosine distance between representations of speech utterances before and after the Adapter module.  Results show that for 73.41\% of the sentences, the Adapter's closest representation, from speech without Adapter set,  was the same sentence, which proves the previous hypothesis that the Adapter module is not mapping the sentences into a completely new space but modifying the sentences created by the speech encoder. The performance improvement from the Adapter may come from disambiguation or additional information provided by the Adapter over the original representation.

The results from these experiments show two main conclusions about the proposed model. Adapter modules do not learn transformation into a new representation space, and the end-to-end training used as baseline fails to capture some useful information for the task and that adding an Adapter between encoder and decoder can be helpful to mitigate it.

Another question about the results is the impact of the trained MultiNMT decoder in the task. The proposed incremental training shows a performance improvement due to transfer learning compared to a randomly initialized decoder when applied to textual data \cite{escolano-etal-2020-talp, escolano2021bilingual}. Results on SLT do not show this improvement and they are similar to the end-to-end baseline without an Adapter. The differences between modalities and obtained representations may explain this behavior. As shown by the analysis on the intermediate representation, adding the Adapter does not show significant improvements. 

\section{Conclusions}
\label{sec:conclusions}

This work presents two main contributions. First, extending a MultiNMT system to perform SLT and zero-shot MultiSLT is possible by coupling language-specific encoder-decoders, even from monolingual ASR data only. Our method eliminates dependencies to MultiSLT data, allowing end-to-end systems and cascade systems to be trained on the same data. Experimental results show that our method provides results on pair with a end-to-end baseline architecture, 0.2 difference for all tested languages, while providing zero-shot SLT even from models trained only on ASR. Second, the Adapter module is a lightweight and effective method to bridge different modalities in an end-to-end model and, on SLT, it can bridge the speech and text representations leading to consistent improvements in all tested translation directions. These improvements are up to +6 BLEU points on the English-to-French SLT direction and up to +1 BLEU points on state-of-the-art end-to-end baseline systems. These improvements are reducing the gap between cascade and end-to-end systems and they are consistent on all zero-shot translation directions for systems trained in SLT tasks.

Analysis of the cross-lingual mappings show that this technique improves the overall performance of the model while maintaining the structure of the representation space, making it suitable for the end-to-end baseline systems and other similar tasks.

Further work includes jointly training speech and text language-specific encoder-decoders and new language addition methods to improve the knowledge transfer from the MultiNMT model to the SLT tasks. Taking advantage of the trained MultiNMT system in both source and target size may lead to better cross-modal mappings and better overall performance. %{\color{red} In addition of  bridging methods between the modalities that have focus more on the mapping of the spaces created between the two modalities.}

\section{Acknowledgements}

Part of this work is done when Carlos Escolano was taking an internship in Telefonica Research. This work is supported by the European Research Council (ERC) under the European Union’s Horizon 2020 research and innovation programme (grant agreement No. 947657.

%\subsection{Set of experiments}

\bibliographystyle{IEEEbib.bst}
\bibliography{interspeech}

% \begin{thebibliography}{9}
% \bibitem[1]{Davis80-COP}
%   S.\ B.\ Davis and P.\ Mermelstein,
%   ``Comparison of parametric representation for monosyllabic word recognition in continuously spoken sentences,''
%   \textit{IEEE Transactions on Acoustics, Speech and Signal Processing}, vol.~28, no.~4, pp.~357--366, 1980.
% \bibitem[2]{Rabiner89-ATO}
%   L.\ R.\ Rabiner,
%   ``A tutorial on hidden Markov models and selected applications in speech recognition,''
%   \textit{Proceedings of the IEEE}, vol.~77, no.~2, pp.~257-286, 1989.
% \bibitem[3]{Hastie09-TEO}
%   T.\ Hastie, R.\ Tibshirani, and J.\ Friedman,
%   \textit{The Elements of Statistical Learning -- Data Mining, Inference, and Prediction}.
%   New York: Springer, 2009.
% \bibitem[4]{YourName17-XXX}
%   F.\ Lastname1, F.\ Lastname2, and F.\ Lastname3,
%   ``Title of your INTERSPEECH 2021 publication,''
%   in \textit{Interspeech 2021 -- 20\textsuperscript{th} Annual Conference of the International Speech Communication Association, September 15-19, Graz, Austria, Proceedings, Proceedings}, 2020, pp.~100--104.
% \end{thebibliography}

\end{document}